\documentclass[11pt,a4paper]{article}
\usepackage[hyperref]{acl2019}
\usepackage{times}
\usepackage{latexsym}

\usepackage{url}

\usepackage{graphicx}
\aclfinalcopy 
\def\aclpaperid{2714} 

\title{Emotion-Cause Pair Extraction: \\A New Task to Emotion Analysis in Texts}

\author{
Rui Xia, Zixiang Ding \\
School of Computer Science and Engineering, \\Nanjing University of Science and Technology, China\\
\texttt{\{rxia, dingzixiang\}@njust.edu.cn}\\
}

\date{}

\begin{document}
\maketitle
\begin{abstract}
  Emotion cause extraction (ECE), the task aimed at extracting the potential causes behind certain emotions in text, has gained much attention in recent years due to its wide applications. However, it suffers from two shortcomings: 1) the emotion must be annotated before cause extraction in ECE, which greatly limits its applications in real-world scenarios; 2) the way to first annotate emotion and then extract the cause ignores the fact that they are mutually indicative. In this work, we propose a new task: emotion-cause pair extraction (ECPE), which aims to extract the potential pairs of emotions and corresponding causes in a document. We propose a 2-step approach to address this new ECPE task, which first performs individual emotion extraction and cause extraction via multi-task learning, and then conduct emotion-cause pairing and filtering. The experimental results on a benchmark emotion cause corpus prove the feasibility of the ECPE task as well as the effectiveness of our approach.
\end{abstract}

\section{Introduction}

Emotion cause extraction (ECE) aims at extracting potential causes that lead to emotion expressions in text.  The ECE task was first proposed and defined as a word-level sequence labeling problem in \citet{lee2010text}. To solve the shortcoming of extracting causes at word level, \citet{gui2016event} released a new corpus which has received much attention in the following study and become a benchmark dataset for ECE research.

\begin{figure*}[!t]
	\centering
	\includegraphics[width=13cm ]{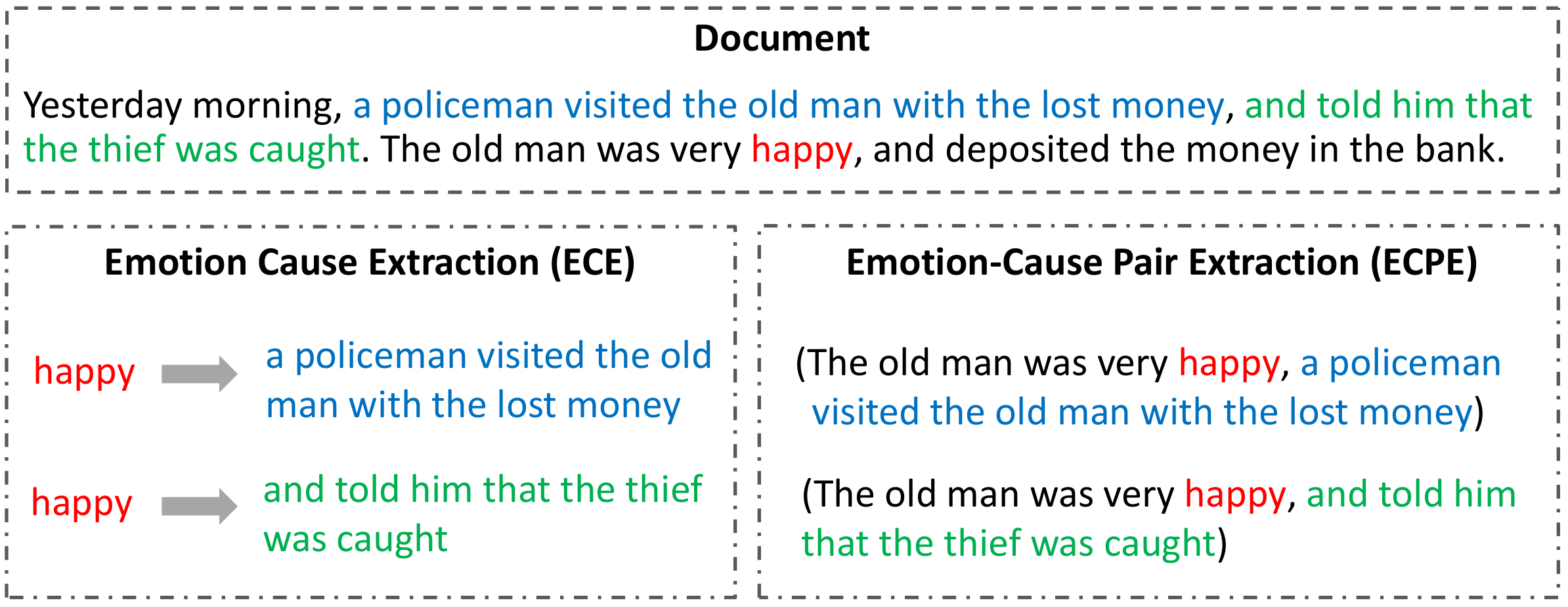}
	\caption{An example showing the difference between the ECE task and the ECPE task.}
	\label{FigureOne}
\end{figure*}

Figure~\ref{FigureOne} displays an example from this corpus, There are five clauses in a document. The emotion ``happy" is contained in the fourth clause. We denote this clause as \textit{emotion clause}, which refers to a clause that contains emotions. It has two corresponding causes: ``a policeman visited the old man with the lost money" in the second clause, and ``told him that the thief was caught" in the third clause. We denote them as \textit{cause clause}, which refers to a clause that contains causes.

The ECE task was formalized as a clause-level binary classification problem in \citet{gui2016event}. The goal is to detect for each clause in a document, whether this clause is a cause given the annotation of emotion. This framework was followed by most of the recent studies in this field \cite{lee2010text,gui2016event,li2018co,xu2019extracting,yu2019multiple}.

However, there are two shortcomings in the current ECE task. The first is that emotions must be annotated before cause extraction in the test set, which limits the applications of ECE in real-world scenarios. The second is that the way to first annotate the emotion and then extract the cause ignores the fact that emotions and causes are mutually indicative.

In this work, we propose a new task: emotion-cause pair extraction (ECPE), which aims to extract all potential pairs of emotions and corresponding causes in a document. In Figure 1 we show the difference between the traditional ECE task and our new ECPE task. The goal of ECE is to extract the corresponding cause clause of the given emotion. In addition to a document as the input, ECE needs to provide annotated emotion at first before cause extraction. In contrast, the output of our ECPE task is a pair of emotion-cause, without the need of providing emotion annotation in advance. Take Figure 1 for example, given the annotation of emotion: ``happy", the goal of ECE is to track the two corresponding cause clauses: ``a policeman visited the old man with the lost money" and ``and told him that the thief was caught". While in the ECPE task, the goal is to directly extract all pairs of emotion clause and cause clause, including (``The old man was very happy", ``a policeman visited the old man with the lost money") and (``The old man was very happy", ``and told him that the thief was caught"), without providing the emotion annotation ``happy".

To address this new ECPE task, we propose a two-step framework. Step 1 converts the emotion-cause pair extraction task to two individual sub-tasks (emotion extraction and cause extraction respectively) via two kinds of multi-task learning networks, with the goal to extract a set of emotion clauses and a set of cause clauses. Step 2 performs emotion-cause pairing and filtering. We combine all the elements of the two sets into pairs and finally train a filter to eliminate the pairs that do not contain a causal relationship.

We evaluated our approach based on a benchmark emotion cause dataset \cite{gui2016event} without using emotion annotations on the test data. We finally achieve the F1 score of 61.28\% in emotion-cause pair extraction.  The experimental results prove the feasibility of the ECPE task and the effectiveness of our approach.

In addition to the emotion-cause pair extraction evaluation, we also evaluate the performance on two individual tasks (emotion extraction and cause extraction).  Without relying on the emotion annotations on the test set, our approach achieves comparable cause extraction performance to traditional ECE methods (slightly lower than the state-of-the-art). In comparison with the traditional ECE methods that removes the emotion annotation dependence, our approach shows great advantages.

The main contributions of this work can be summarized as follows:

\begin{itemize}
	\item We propose a new task: emotion-cause pair extraction (ECPE). It solves the shortcomings of the traditional ECE task that depends on the annotation of emotion before extracting cause, and allows emotion cause analysis to be applied to real-world scenarios.
	\item We propose a two-step framework to address the ECPE task, which first performs individual emotion extraction and cause extraction and then conduct emotion-cause pairing and filtering. 
	\item Based on a benchmark ECE corpus, we construct a corpus suitable for the ECPE task. The experimental results prove the feasibility of the ECPE task as well as the effectiveness of our approach.
\end{itemize}


\section{Related Work}
\citet{lee2010text} first presented the task of emotion cause extraction (ECE) and defined this task as extracting the word-level causes that lead to the given emotions in text. They constructed a small-scale Chinese emotion cause corpus in which the spans of both emotion and cause were annotated. Based on the same task settings, there were some other individual studies that conducted ECE research on their own corpus using rule based methods \cite{neviarouskaya2013extracting, li2014text, gao2015emotion, gao2015rule, yada2017bootstrap} or machine learning methods \cite{ghazi2015detecting, song2015detecting}.

\citet{chen2010emotion} suggested that a clause may be the most appropriate unit to detect causes based on the analysis of the corpus in \cite{lee2010text}, and transformed the task from word-level to clause-level. They proposed a multi-label approach that detects multi-clause causes and captures the long-distance information. There were a lot of work based on this task setting. \citet{russo2011emocause} introduced a method based on the linguistic patterns and common sense knowledge for the identification of Italian sentences which contain a cause phrase. \citet{gui2014emotion} used 25 manually complied rules as features, and chose machine learning models, such as SVM and CRFs, to detect causes. \citet{gui2016event}, \citet{gui2016emotion} and \citet{xu2017ensemble} released a Chinese emotion cause dataset using SINA city news. This corpus has received much attention in the following study and has become a benchmark dataset for ECE research. Based on this corpus, several traditional machine learning methods \cite{gui2016event, gui2016emotion, xu2017ensemble} and  deep learning methods \cite{gui2017question, li2018co, yu2019multiple, xu2019extracting} were proposed.

In addition, \citet{cheng2017emotion} focused on cause detection for Chinese microblogs using a multiple-user structure. They formalized two cause detection tasks for microblogs (current-subtweet-based cause detection and original-subtweet-based cause detection) and introduced SVM and LSTM to deal with them. \citet{chen2018joint} presented a neural network-based joint approach for emotion classification and cause detection in order to capture mutual benefits across these two sub-tasks. \citet{chen2018hierarchical} proposed a hierarchical Convolution Neural Network (Hier-CNN), which used clause-level encoder and subtweet-level encoder to incorporate the word context features and event-based features respectively.

All of the above work attempts to extract word-level or clause-level causes given the emotion annotations.  While our work is different from them, we propose to extract both the emotion and the corresponding causes at the same time (i.e., emotion-cause pair extraction) and to investigate whether indicating causes can improve emotion extraction and vice versa. Since we believe that cause and emotion are not mutually independent.

\section{Task}
First of all, we give the definition of our emotion-cause pair extraction (ECPE) task. Given a document consisting of multiple clauses $ d=[c_1,c_2,...,c_{|d|} ] $,  the goal of ECPE is to extract a set of emotion-cause pairs in $ d $:
\begin{equation}\label{key}
P= \{ \cdots,(c^e,c^c ),\cdots \},
\end{equation}
where $ c^e $ is an emotion clause and $ c^c $ is the corresponding cause clause

In traditional emotion cause extraction task, the goal is to extract $ c^c $ given the annotation of $ c^e: c^e \rightarrow  c^c $. In comparison, the ECPE task is new and more difficult to address, because the annotation of emotion $ c^e $ is not provided before extraction.

Note that similar as the traditional ECE task, the ECPE task is also defined at the clause level, due to the difficulty describing emotion causes at the word/phrase level. It means that the ``emotion" and ``cause" used in this paper refer to ``emotion clause" and ``cause clause" respectively.

\section{Approach}
In this work, we propose a two-step approach to address this new ECPE task:

\begin{itemize}
	\item \textbf{Step 1 (Individual Emotion and Cause Extraction)}. We first convert the emotion-cause pair extraction task to two individual sub-tasks (emotion extraction and cause extraction respectively). Two kinds of multi-task learning networks are proposed to model the two sub-tasks in a unified framework, with the goal to extract a set of emotion clauses $ E=\{c_1^e,\cdots,c_m^e \} $ and a set of cause clauses $ C=\{c_1^c,\cdots,c_n^c \} $ for each document.
	\item \textbf{Step 2 (Emotion-Cause Pairing and Filtering)}. We then pair the emotion set $ E $ and the cause set $ C $ by applying a Cartesian product to them. This yields a set of candidate emotion-cause pairs. We finally train a filter to eliminate the pairs that do not contain a causal relationship between emotion and cause.
	
\end{itemize}

\subsection{Step 1: Individual Emotion and Cause Extraction}
The goal of Step 1 is to extract a set of emotion clauses and a set of cause clauses for each document, respectively. To this end, we propose two kinds of multi-task learning networks, (i.e., Independent Multi-task Learning and Interactive Multi-task Learning). The latter is an enhanced version that further captures the correlation between emotion and cause on the basis of the former.

\subsubsection{Independent Multi-task Learning}
\begin{figure}
	\centering
	\includegraphics[width=1.05\columnwidth ]{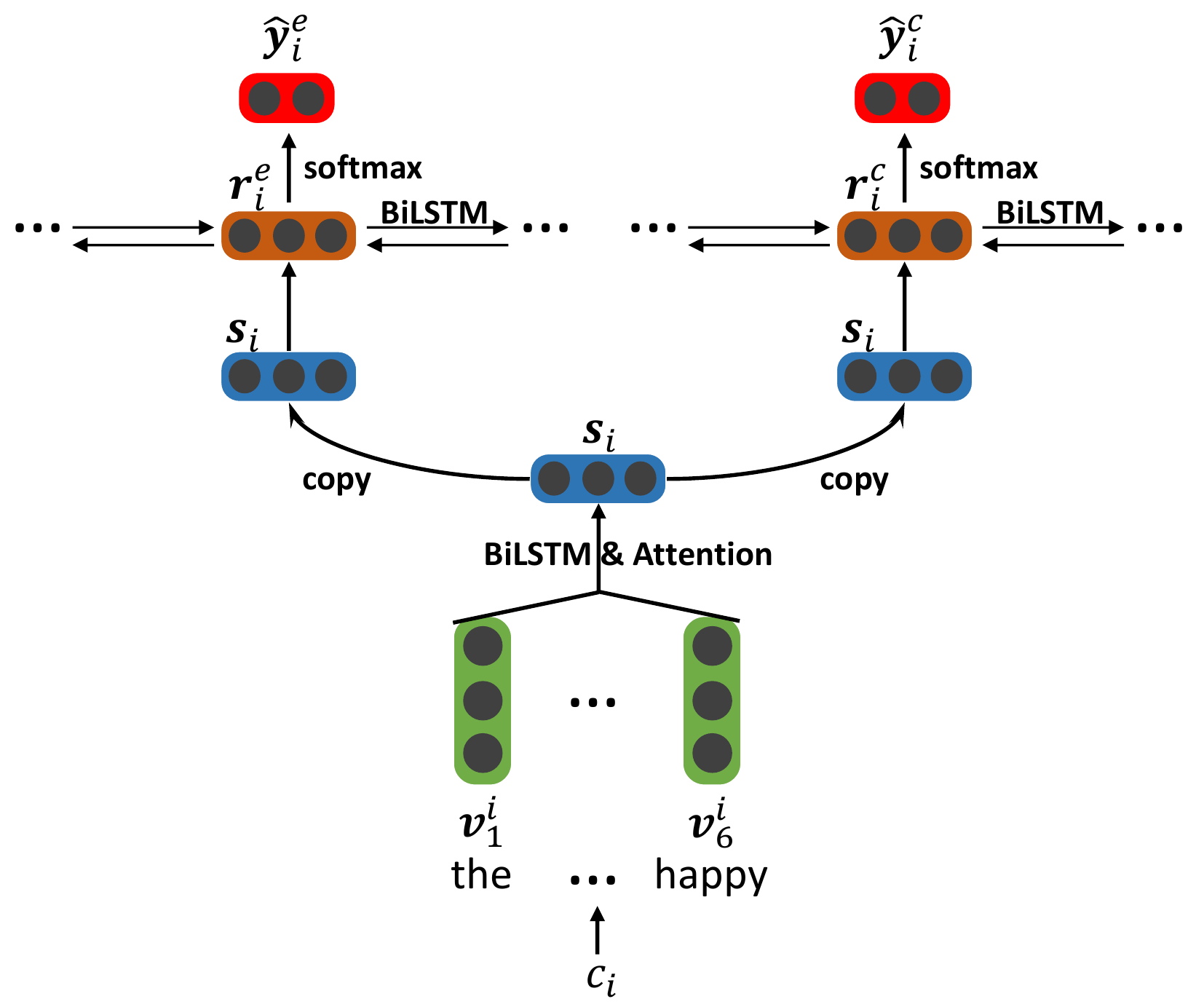}
	\caption{The Model for Independent Multi-task Learning (Indep).}
	\label{FigureTwo}
\end{figure}

In our task, a document contains multiple clauses: $ d=[c_1,c_2,...,c_{|d|})] $, and each $ c_i $ also contains multiple words $ c_i=[w_{i,1},w_{i,2},...,w_{i,|c_i |}] $. To capture such a ``word-clause-document" structure, we employ a Hierarchical Bi-LSTM network which contains two layers, as shown in Figure~\ref{FigureTwo}.

The lower layer consists of a set of word-level Bi-LSTM  modules, each of which corresponds to one clause, and accumulate the context information for each word of the clause. The hidden state of the $ j $th word in the $ i $th clause $ \mathbf{h}_{i,j} $ is obtained based on a bi-directional LSTM. Attention mechanism is then adopt to get a clause representation $ \mathbf{s}_i $. Here we omit the details of Bi-LSTM and attention for limited space, readers can refer to \citet{graves2013speech} and \citet{bahdanau2014neural}.

The upper layer consists of two components: one for emotion extraction and another for cause extraction. Each component is a clause-level Bi-LSTM which receives the independent clause representations $ [\mathbf{s}_1,\mathbf{s}_2,...,\mathbf{s}_{|d|} ] $ obtained at the lower layer as inputs. The hidden states of two component Bi-LSTM,  $ \mathbf{r}_i^e $ and $ \mathbf{r}_i^c $ , can be viewed as the context-aware representation of clause $ c_i $, and finally feed to the softmax layer for emotion prediction and cause predication:
\begin{equation}
\label{eqn_example}
\hat{\mathbf{y}}^e_i = \textrm{softmax}( \mathbf{W}^e\mathbf{r}_i^e + \mathbf{b}^e),
\end{equation}
\begin{equation}
\label{eqn_example}
\hat{\mathbf{y}}^c_i = \textrm{softmax}( \mathbf{W}^c\mathbf{r}_i^c + \mathbf{b}^c),
\end{equation}
where the superscript $ e $ and $ c $ denotes emotion and cause, respectively.

The loss of the model is a weighted sum of two components:
\begin{equation}\label{key}
L^p=\lambda L^e+(1-\lambda) L^c,
\end{equation}
where $ L^e $ and $ L^c $ are the cross-entropy error of emotion predication and cause predication respectively, and $ \lambda $ is a tradeoff parameter.

\subsubsection{Interactive Multi-task Learning}
Till now, two component Bi-LSTM at the upper layer are independent to each other. However, as we have mentioned, the two sub-tasks (emotion extraction and cause extraction) are not mutually independent. On the one hand, providing emotions can help better discover the causes; on the other hand, knowing causes may also help more accurately extract emotions.

\begin{figure*}
	\centering
	\includegraphics[width=13cm ]{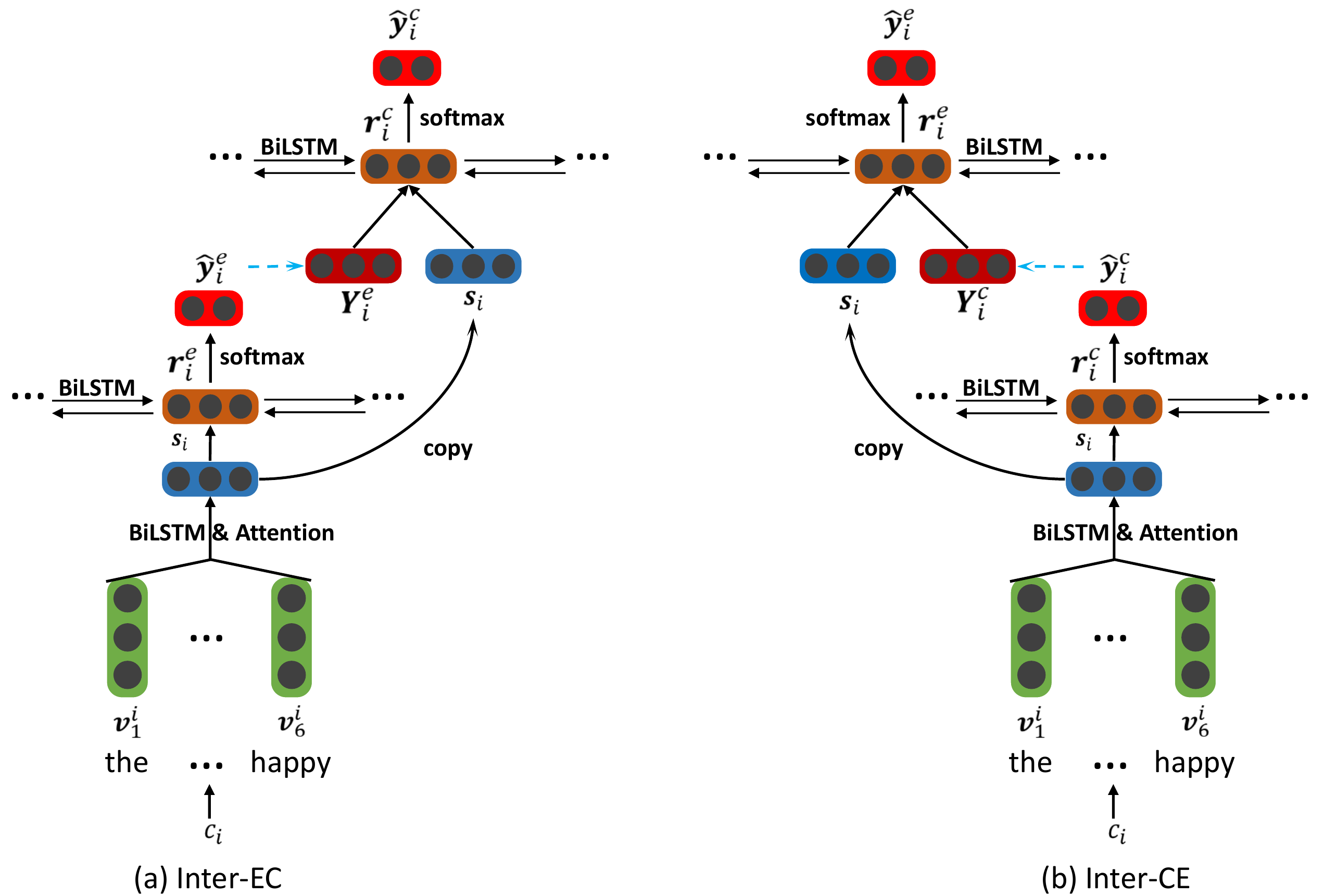}
	\caption{Two Models for Interactive Multi-task Learning: (a) Inter-EC, which uses emotion extraction to improve cause extraction (b) Inter-CE, which uses cause extraction to enhance emotion extraction.}
	\label{FigureThree}
\end{figure*}

Motivated by this, we furthermore propose an interactive multi-task learning network, as an enhanced version of the former one, to capture the correlation between emotion and cause. The structure is shown in Figure~\ref{FigureThree}. It should be noted that the method using emotion extraction to improve cause extraction is called Inter-EC. In addition, we can also use cause extraction to enhance emotion extraction, and call this method Inter-CE. Since Inter-EC and Inter-CE are similar in structure, we only introduce Inter-EC (illustrated in Figure~\ref{FigureThree} (a) ) instead of both.

Compared with Independent Multi-task Learning, the lower layer of Inter-EC is unchanged, and the upper layer consists of two components, which are used to make predictions for emotion extraction task and cause extraction task in an interactive manner. Each component is a clause-level Bi-LSTM followed by a softmax layer.

The first component takes the independent clause representations  $ [\mathbf{s}_1,\mathbf{s}_2,...,\mathbf{s}_{|d|} ] $ obtained at the lower layer as inputs for emotion extraction. The hidden state of clause-level Bi-LSTM  $ \mathbf{r}_i^{e} $ is used as feature to predict the distribution of the $ i $-th clause  $ \hat{\mathbf{y}}_i^{e} $. Then we embed the predicted label of the $ i $-th clause as a vector  $ \mathbf{Y}_i^{e} $, which is used for the next component.

Another component takes $ ( \mathbf{s}_1 \oplus \mathbf{Y}^{e}_1, \mathbf{s}_2 \oplus \mathbf{Y}^{e}_2,..., \mathbf{s}_{|d|} \oplus \mathbf{Y}^{e}_{|d|}  ) $ as inputs for cause extraction, where $ \oplus $ represents the concatenation operation. The hidden state of clause-level Bi-LSTM $ \mathbf{r}_i^{c} $ is used as feature to predict the distribution of the $ i $-th clause  $ \hat{\mathbf{y}}_i^{c} $.

The loss of the model is a weighted sum of two components, which is the same as Equation 4.

\subsection{Step 2: Emotion-Cause Pairing and Filtering}
In Step 1, we finally obtain a set of emotions $ E=\{c_1^e,\cdots,c_m^e \} $ and a set of cause clauses $ C=\{c_1^c,\cdots,c_n^c \} $ . The goal of Step 2 is then to pair the two sets and construct a set of emotion-cause pairs with causal relationship.

Firstly, we apply a Cartesian product to $ E $ and $ C $, and obtain the set of all possible pairs:
\begin{equation}\label{key}
P_{all}= \{ \cdots,(c^e_i,c^c_j ),\cdots \},
\end{equation}

Secondly, we represent each pair in $ P_{all} $ by a feature vector composed of three kinds of features:

\begin{equation}\label{key}
\mathbf{x}_{(c^e_i,c^c_j )} = [\mathbf{s}^e_i, \mathbf{s}^c_j, \mathbf{v}^d],
\end{equation}
where  $ \mathbf{s}^e $ and $ \mathbf{s}^c $ are the representations of the emotion clause and cause clause respectively, and $ \mathbf{v}^d $ represents the distances between the two clauses.

A Logistic regression model is then trained to detect for each candidate pair $ (c_i^e,c_j^c ) $, whether $ c_i^e $ and $ c_j^c $ have a causal relationship:

\begin{equation}\label{key}
\hat{y}_{(c^e_i,c^c_j )} \leftarrow \delta (\mathbf{\theta}^{\rm T} \mathbf{x}_{(c^e_i,c^c_j )}),
\end{equation}
where $ \hat{y}_{(c^e_i,c^c_j )}=1 $ denotes that $ (c^e_i,c^c_j ) $ is a pair with causal relationship,  $ \hat{y}_{(c^e_i,c^c_j )}=0 $ denotes $ (c^e_i,c^c_j ) $ is a pair without causal relationship, and $ \delta(\cdot) $ is the Sigmoid function.
We finally remove the pairs whose $ \hat{y}_{(c^e_i,c^c_j )} $ is 0 from $ P_{all} $, and get the final set of emotion-cause pairs.

\begin{table*}
	\small
	\centering
	\begin{tabular} {c|c|c}
		\hline   & Number & Percentage\\
		\hline
		Documents with one emotion-cause pair & 1746 & 89.77\% \\
		Documents with two emotion-cause pairs &	177 &	9.10\% \\
		Documents with more than two emotion-cause pairs &	22 &	1.13\% \\
		All &	1945 &	100\% \\
		\hline
	\end{tabular}
	\caption{The proportion of documents with different number of emotion-cause pairs in the merged dataset. }
	\label{TableTwo}
\end{table*}

\section{Experiments}

\subsection{Dataset and Metrics}
Since there was no directly available corpus for the ECPE task, we constructed a ECPE corpus based on the benchmark ECE corpus  \cite{gui2016event}, in which each document contains only one emotion and corresponding one or more causes. Documents having two or more emotions are split into several samples such that each contains only one emotion. In order to better meet the ECPE task settings, we merged the documents with the same text content into one document, and labeled each {emotion, cause} pair in this document. The proportion of documents with different number of emotion-cause pairs in the combined dataset are shown in Table~\ref{TableTwo}.

\begin{table*}
	\small
	\centering
	
	\begin{tabular} {c|c|c|c|c|c|c|c|c|c}
		\hline
		  & \multicolumn{3}{|c}{emotion extraction} & \multicolumn{3}{|c}{cause extraction} & \multicolumn{3}{|c}{emotion-cause pair extraction}\\
		\hline
		& $P$ & $R$ & $F1$ & $P$ & $R$ & $F1$ & $P$ & $R$ & $F1$ \\
		\hline
		\textbf{Indep} & 0.8375 & 0.8071 & 0.8210 & 0.6902 & 0.5673 & 0.6205 & 0.6832 & 0.5082 & 0.5818 \\
		\textbf{Inter-CE} & \textbf{0.8494} & \textbf{0.8122} & \textbf{0.8300} & 0.6809 & 0.5634 & 0.6151 & \textbf{0.6902} & 0.5135 & 0.5901\\
		\textbf{Inter-EC} & 0.8364 & 0.8107 & 0.8230 & \textbf{0.7041} & \textbf{0.6083} & \textbf{0.6507} & 0.6721 & \textbf{0.5705} & \textbf{0.6128}\\
		
		\hline
	\end{tabular}
	\caption{Experimental results of all proposed models and variants using precision, recall, and F1-measure as metrics on the ECPE task as well as the two sub-tasks.}
	\label{TableThree}
\end{table*}

\begin{table*}
	\small
	\centering
	
	\begin{tabular} {c|c|c|c|c|c|c|c|c|c}
		\hline
		& \multicolumn{3}{|c}{emotion extraction} & \multicolumn{3}{|c}{cause extraction} & \multicolumn{3}{|c}{emotion-cause pair extraction}\\
		\hline
		& $P$ & $R$ & $F1$ & $P$ & $R$ & $F1$ & $P$ & $R$ & $F1$ \\
		\hline
		\textbf{Inter-CE-Bound} & \#0.9144 & \#0.8894 & \#0.9016 & \#1.0000 & \#1.0000 & \#1.0000 & \#0.8682 & \#0.8806 & \#0.8742\\
		\textbf{Inter-EC-Bound} & \#1.0000 & \#1.0000 & \#1.0000 & \#0.7842 & \#0.7116 & \#0.7452 & \#0.7610 & \#0.7084 & \#0.7328\\
		
		\hline
	\end{tabular}
	\caption{Results of upperbound experiments for Inter-CE and Inter-EC.}
	\label{TableThree2}
\end{table*}

We stochastically select 90\% of the data for training and the remaining 10\% for testing. In order to obtain statistically credible results, we repeat the experiments 20 times and report the average result. We use the precision, recall, and F1 score as the metrics for evaluation, which are calculated as follows:
\begin{equation}
\label{eqn_example}
P= \frac {\sum correct\_pairs} {\sum proposed\_pairs},
\end{equation}
\begin{equation}
\label{eqn_example}
R= \frac {\sum correct\_pairs} {\sum annotated\_pairs},
\end{equation}
\begin{equation}
\label{eqn_example}
F1= \frac {2 \times P \times R} { P + R} ,
\end{equation}
where $ proposed\_pairs $ denotes the number of emotion-cause pairs predicted by the model, $ annotated\_pairs $ denotes the total number of emotion-cause pairs that are labeled in the dataset and the $ correct\_pairs $ means the number of pairs that are both labeled and predicted as an emotion-cause pair.

In addition, we also evaluate the performance of two sub-tasks: emotion extraction and cause extraction. The precision, recall and F1 score defined in \citet{gui2016event} are used as the evaluation metrics.

\begin{table*}
	\small
	\centering
	
	\begin{tabular} {c|c|c|c|c|c|c|c}
		\hline
		  & \multicolumn{3}{|c|}{without emotion-cause pair filtering} & \multicolumn{4}{|c}{with emotion-cause pair filtering}\\
		\hline
		& $P$ & $R$ & $F1$ & $P$ & $R$ & $F1$ & $keep\_rate$ \\
		\hline
		\textbf{Indep} & 0.5894 & 0.5114 & 0.5451 & 0.6832 & 0.5082 & 0.5818 & 0.8507\\
		\textbf{Inter-CE} & 0.5883 & 0.5192 & 0.5500 & \textbf{0.6902} & 0.5135 & 0.5901 & 0.8412\\
		\textbf{Inter-EC} & \textbf{0.6019} & \textbf{0.5775} & \textbf{0.5842} & 0.6721 & \textbf{0.5705} & \textbf{0.6128} & 0.8889\\

		\textbf{Inter-CE-Bound} & \#0.8116 & \#0.8880  & \#0.8477 & \#0.8682 & \#0.8806 & \#0.8742 & 0.9271\\
		\textbf{Inter-EC-Bound} & \#0.6941 & \#0.7118 & \#0.7018 & \#0.7610 & \#0.7084 & \#0.7328 & 0.9088\\
		
		\hline
	\end{tabular}
	\caption{Experimental results of all proposed models and variants using precision, recall, and F1-measure as metrics on the ECPE task with or without the pair filter.}
	\label{TableFour}
\end{table*}

\subsection{Experimental Settings}
We use word vectors that were pre-trained on the corpora from Chinese Weibo\footnote{http://www.aihuang.org/p/challenge.html} with word2vec \cite{mikolov2013distributed} toolkit. The dimension of word embedding is set to 200. The number of hidden units in BiLSTM for all our models is set to 100. All weight matrices and bias are randomly initialized by a uniform distribution U($ -0.01 $, $ 0.01 $).

For training details, we use the stochastic gradient descent (SGD) algorithm and Adam update rule with shuffled minibatch. Batch size and learning rate are set to 32 and 0.005, respectively. As for regularization, dropout is applied for word embeddings and the dropout rate is set to 0.8. Besides, we perform L2 constraints over the soft-max parameters and L2-norm regularization is set as 1e-5.\footnote{The source code and merged corpus can be obtained at https://github.com/NUSTM/ECPE}

\subsection{Evaluation on the ECPE Task}

\noindent \textbf{(1) Overall Performance}

In Table~\ref{TableThree}, we report the experimental results of the following three proposed models on three tasks (emotion extraction, cause extraction and emotion-cause pair extraction).

\begin{itemize}
	\item \textbf{Indep}: Indep denotes the method proposed in section 4.1.1. In this method, emotion extraction and cause extraction are independently modeled by two Bi-LSTMs.
	\item \textbf{Inter-CE}: Inter-CE denotes the method proposed in section 4.1.2, where the predictions of cause extraction are used to improve emotion extraction.
	\item \textbf{Inter-EC}: Inter-EC denotes the method proposed in section 4.1.2, where the predictions of emotion extraction are used to enhance cause extraction.
\end{itemize}

Compared with Indep, Inter-EC gets great improvements on the ECPE task as well as the two sub-tasks. Specifically, we find that the improvements are mainly in the recall rate on the cause extraction task, which finally lead to the great improvement in the recall rate of ECPE. This shows that the predictions of emotion extraction are helpful to cause extraction and proves the effectiveness of Inter-EC. In addition, the performance of emotion extraction also improved, which indicates that the supervision from cause extraction is also beneficial for emotion extraction. 

Inter-CE also gets significant improvements on the ECPE task compared to Indep. Specifically, we find that the improvements are mainly in the precision score on the emotion extraction task, which finally lead to the significant improvement in the precision score of ECPE. This shows that the predictions of cause extraction are beneficial to emotion extraction and proves the effectiveness of Inter-CE.  

By comparing Inter-EC and Inter-CE, we find that the improvement of Inter-EC is mainly obtained on the cause extraction task, and the improvement of Inter-CE is mainly gained on the emotion extraction task. These results are consistent with our intuition that emotion and cause are mutually indicative. In addition, we find that the improvements of Inter-EC on the cause extraction task are much more than the improvement of Inter-CE on the emotion extraction task. We guess that it is because cause extraction is more difficult than emotion extraction, hence there is more room for extra improvement.

\vspace{2mm}

\noindent \textbf{(2) Upper-Bound of Emotion and Cause Interaction}

In order to further explore the effect of sharing predictions of two sub-tasks, we designed upperbound experiments for Inter-CE and Inter-EC. The results are shown in Table~\ref{TableThree2}.

\begin{itemize}
	\item \textbf{Inter-CE-Bound}: Inter-CE-Bound is a variant of Inter-CE that uses the label of cause extraction to help emotion extraction.
	\item \textbf{Inter-EC-Bound}: Inter-EC-Bound is a variant of Inter-EC that uses the label of emotion extraction to help cause extraction. 	
\end{itemize}

The results of Inter-CE-Bound and Inter-EC-Bound are preceded by a ``\#", indicating that they cannot be compared fairly with other methods because they use annotations. Compared with Indep, the performance of Inter-EC-Bound on cause extraction and the performance of Inter-CE-Bound on emotion extraction both improve greatly. Moreover, the improvement of Inter-EC-Bound on the cause extraction task are much more than the improvement of Inter-CE-Bound on the emotion extraction task. We guess this is because the cause extraction task is more difficult than the emotion extraction task, and there is more room for improvement, which is consistent with previous section.

By comparing the results of Inter-EC-Bound and Inter-EC, we found that although Inter-EC performs better than Indep, it is far poorer than Inter-EC-Bound, which is caused by lots of errors in the predictions of emotion extraction. We can draw the same conclusion when comparing Inter-CE-Bound and Inter-CE.

These experimental results further illustrate that emotion and cause are mutually indicative, and indicate that if we can improve the performance of emotion extraction task, we can get better performance on cause extraction task and vice versa, which finally lead to the improvement on ECPE. But it should be noted it is only an upper-bound experiment where the ground-truth of emotion/causes are used to predict each other.

\vspace{2mm}

\noindent \textbf{(3) Effect of Emotion-Cause Pair Filtering}

In Table~\ref{TableFour}, we report the emotion-cause pair extraction performance with/without pair filtering. With/Without pair filtering indicates whether we adopt a pair filter after applying a Cartesian product in the second step. $keep\_rate$ indicates the proportion of emotion-cause pairs in $ P_{all} $ that are finally retained after pair filtering.

An obvious observation is that the F1 scores of all models on the ECPE task are significantly improved by adopting the pair filter. These results demonstrate the effectiveness of the pair filter. Specifically, by introducing the pair filter, some of the candidate emotion-cause pairs in $ P_{all} $ are filtered out, which may result in a decrease in the recall rate and an increase in precision. According to Table~\ref{TableFour}, the precision scores of almost all models are greatly improved (more than 7\%), in contrast, the recall rates drop very little (less than 1\%), which lead to the significant improvement in F1 score.

\subsection{Evaluation on the ECE task}
In Table~\ref{TableFive}, we further examine our approach by comparing it with some existing approaches on the traditional ECE task. It should be noted that our Inter-EC model does not use the emotion annotations on the test data.

\begin{table}
	\small
	\centering
	\begin{tabular} {c|c|c|c}
		\hline
		 & $P$ & $R$ & $F1$\\
		\hline
		\textbf{RB} & 0.6747 & 0.4287 & 0.5243\\
		\textbf{CB} & 0.2672 & 0.7130 & 0.3887\\
		\textbf{RB+CB+ML} & 0.5921 & 0.5307 & 0.5597\\
		\textbf{Multi-Kernel} & 0.6588 & 0.6927 & 0.6752\\
		\textbf{Memnet} & 0.5922 & 0.6354 & 0.6134\\
		\textbf{ConvMS-Memnet} & 0.7076 & 0.6838 & 0.6955\\
		\textbf{CANN} & 07721 & 0.6891 & 0.7266\\
		\hline
		\textbf{CANN-E} & 0.4826 & 0.3160 & 0.3797\\
		\textbf{Inter-EC} & 0.7041 & 0.6083 & 0.6507\\
		\hline
	\end{tabular}
	\caption{Experimental results of some existing ECE approaches and our model on the ECE task.}
	\label{TableFive}
\end{table}
\begin{itemize}
	\item \textbf{RB} is a rule-based method with manually defined linguistic rules \cite{lee2010text}.
	\item \textbf{CB} is a method based on common-sense knowledge \cite{russo2011emocause}.
	\item \textbf{RB+CB+ML} (Machine learning method trained from rule-based features and common-sense knowledge base) uses rules and facts in a knowledge base as features and a traditional SVM classifier for classification \cite{chen2010emotion}.
	\item \textbf{Multi-kernel} uses the multi-kernel method to identify the cause \cite{gui2016event}.
	\item \textbf{Memnet} denotes a deep memory network proposed by \citet{gui2017question}.
	\item \textbf{ConvMS-Memnet} is a convolutional multiple-slot deep memory network proposed by \citet{gui2017question}.
	\item \textbf{CANN} denotes a co-attention neural network model proposed in \citet{li2018co}.
\end{itemize}

It can be seen that although our method does not use emotion annotations on the test data, it still achieves comparable results with most of the traditional methods for the ECE task. This indicates that our method can overcome the limitation that emotion annotations must be given at the testing phase in the traditional ECE task, but without reducing the cause extraction performance.

In order to compare with the traditional methods for the ECE task under the same experimental settings, we furthermore implemented a simplification of CANN (CANN-E), which removes the dependency of emotion annotation in the test data. 

It is clear that by removing the emotion annotations, the F1 score of CANN drops dramatically (about 34.69\%). In contrast, our method does not need the emotion annotations and achieve 65.07\% in F1 measure, which significantly outperforms the CANN-E model by 27.1\%.

\section{Conclusions and Future Work}
In this paper, we propose a new task: emotion-cause pair extraction, which aims to extract potential pairs of emotions and corresponding causes in text. To deal with this task, we propose a two-step method, in which we first extract both emotions and causes respectively by multi-task learning, then combine them into pairs by applying Cartesian product, and finally employ a filter to eliminate the false emotion-cause pairs. Based on a benchmark ECE corpus, we construct a corpus suitable for the ECPE task. The experimental results prove the effectiveness of our method. 

The two-step strategy may not be a perfect solution to solve the ECPE problem. On the one hand, its goal is not direct. On the other hand, the mistakes made in the first step will affect the results of the second step. In the future work, we will try to build a one-step model that directly extract the emotion-cause pairs in an end-to-end fashion.

\section*{Acknowledgments}
\noindent The work was supported by the Natural Science Foundation of China (No. 61672288), and the Natural Science Foundation of Jiangsu Province for Excellent Young Scholars (No. BK20160085). Rui Xia and Zixiang Ding contributed equally to this paper.

\bibliography{acl2019_camera_ready}
\bibliographystyle{acl_natbib}

\end{document}